\documentclass[10pt,a4paper]{article}
\usepackage[utf8]{inputenc}
\usepackage[english]{babel}
\usepackage[babel]{csquotes}
\usepackage{amsmath}
\usepackage{mathtools}
\usepackage{authblk}
\usepackage{float}
\usepackage{titling}
\usepackage{caption}
\usepackage{blindtext}
\usepackage{graphicx}
\graphicspath{ {C:\Users\treppner\bwSyncAndShare\Fitness_Tracker} }
\usepackage{amsfonts}
\usepackage{amssymb}
\usepackage[round]{natbib}
\usepackage{hyperref}
\author{Martin Treppner}
\title{Modeling Activity Tracker Data Using Deep Boltzmann Machines}
\begin{document}
\setlength{\parindent}{0em}

\begin{titlingpage}
\title{Modeling Activity Tracker Data Using Deep Boltzmann Machines}
\date{\today}

\author{Stefan Lenz}
\author{Harald Binder}
\author{Daniela Zöller}
\affil{Institute of Medical Biometry and Statistics,\\
Faculty of Medicine and Medical Center - University of Freiburg}
\maketitle

\begin{abstract}
	Commercial activity trackers are set to become an essential tool in health research, due to increasing availability in the general population. The corresponding vast amounts of mostly unlabeled data pose a challenge to statistical modeling approaches. To investigate the feasibility of deep learning approaches for unsupervised learning with such data, we examine weekly usage patterns of Fitbit activity trackers with deep Boltzmann machines (DBMs). This method is particularly suitable for modeling complex joint distributions via latent variables. We also chose this specific procedure because it is a generative approach, i.e., artificial samples can be generated to explore the learned structure. We describe how the data can be preprocessed to be compatible with binary DBMs. The results reveal two distinct usage patterns in which one group frequently uses trackers on Mondays and Tuesdays, whereas the other uses trackers during the entire week. This exemplary result shows that DBMs are feasible and can be useful for modeling activity tracker data.
\end{abstract}
\end{titlingpage}

\newpage

\section{Introduction}

Wearable devices, more specifically activity trackers, are attracting considerable interest due to their ability to measure physical activity continuously. The identification of patterns in the corresponding vast amounts of data generated in these settings poses a challenge to conventional linear modeling approaches.

Few researchers have addressed the problem of modeling the joint distribution of large quantities of activity tracker data using deep learning techniques. This paper seeks to address this issue by employing a deep Boltzmann machine (DBM) which has been shown to be a promising method in many applications like single-cell genomics, object recognition, and information retrieval  \citep{angermueller2017accurate,pmlr-v5-salakhutdinov09a,DBLP:journals/corr/SrivastavaSH13}. Additionally, it provides a generative model, i.e., artificial samples can be generated from a trained model for exploring the structure learned by the approach.
Our analysis aims to develop methods that can detect structures in large unlabeled datasets and broaden current knowledge of weekly activity tracker usage patterns. Furthermore, the results could subsequently provide information on associations of latent activity patterns with health outcomes.

Due to the complex structure of activity tracker data, the modeling task is often subdivided. \cite{bai2017two}, for example, employed a two-stage model in which the presence of activity and the activity intensity given any activity at all are modeled separately using linear models. In \cite{ellis2014random}, the authors utilized random forests to model physical activity but were only able to train their model on data from a strictly controlled setting, which is not directly comparable to activity data from the use of activity trackers by the general population. In contrast to the approach considered in our study, \cite{staudenmayer2009artificial} use supervised learning methods in a controlled surrounding, i.e., a gold standard is needed in the learning procedure.

\section{Data preprocessing}

The data for this study was obtained from \emph{openhumans.org} where users of activity trackers can donate their activity records for the purpose of scientific studies. We had access to publicly shared Fitbit data from 29 individuals and extracted their daily step counts specifically. Since our goal was to explore weekly usage patterns data was prepared accordingly. Hence, we define $\textbf{X}_{ij}$ to be the recorded number of steps in a given week $i$ at day $j$,$j = 1$ corresponding to Monday, and so on.

After deleting all weeks in which no data was recorded, we dichotomized the step counts for activity/inactivity of the tracker. This approach was adapted by \cite{bai2017two} using an indicator function:

\begin{equation}
\mathbf{1}_A (x_{ij}) \coloneqq \begin{cases} 1 &, x_{ij} > 0 \\ 0 &, else \end{cases}
\end{equation} 

\section{Methods}

We chose the framework of a deep Boltzmann machine because it is one of the most practical ways to learn large joint distributions while still being able to perform inference tasks. The following subsections give a brief overview of deep Boltzmann machines and how these models can be trained efficiently

\subsection{Deep Boltzmann Machines (DBMs)}

In order to identify the usage patterns mentioned above, we employ deep Boltzmann machines. This method has the potential to outperform previous approaches for wearable device data because it can learn powerful representations of complex joint distributions \cite[p.3173]{hess2017partitioned}. In addition, DBMs can process vast quantities of unlabeled data which is inevitable in settings where data is obtained from fitness trackers under real-life conditions.

In our approach, we consider a two-layer Boltzmann machine where we denote the visible layer as $\textbf{v}$ and use $\textbf{h}^{1}$,$\textbf{h}^{2}$ for the first and second hidden layer, respectively. Furthermore, we restrict ourselfes to a DBM with no within layer connections. The DBM enables modeling of the joint distribution of a large number of Bernoulli variables. In this context, these variables represent whether an individual made use of an activity tracker or not at a specific day. Broadly speaking, DBMs consist of stacked sets of visible and hidden nodes in which each layer captures complex, higher-order correlations between the activities of hidden features in the layer below \cite[p.450]{pmlr-v5-salakhutdinov09a}.

Following the definitions and outline of \cite{pmlr-v5-salakhutdinov09a} we define the energy of the state $ \lbrace \textbf{v}, \textbf{h}^{1}, \textbf{h}^{2} \rbrace $ as:

\begin{equation}
E(\textbf{v},\textbf{h}^{1},\textbf{h}^{2};\theta) = -\textbf{v}^{T}\textbf{W}^{1}\textbf{h}^{1} -\textbf{h}^{1T}\textbf{W}^{2}\textbf{h}^{2}
\end{equation}

where $\theta = \lbrace \textbf{W}^{1},\textbf{W}^{2}\rbrace $ are the model parameters, representing the symmetric interactions between layers. In energy-based models, low energy corresponds to high probabilities whereas high energy represents a low probability.

Next, we define the probability of the visible vector $\textbf{v}$:

\begin{equation}
p(\textbf{v};\theta) = \frac{1}{Z(\theta)}\sum_{\textbf{h}^{1},\textbf{h}^{2}}{exp(-E(\textbf{v},\textbf{h}^{1},\textbf{h}^{2};\theta))}
\end{equation}

Furthermore, the conditional distributions over the visible and the two sets of hidden units are given by logistic functions $\sigma$ : 

\begin{equation}
p(h_{j}^{1} = 1 | \textbf{v},\textbf{h}^{2}) = \sigma \left( \sum_{i}{W_{ij}^{1}v_{i}} + \sum_{m}{W_{jm}^{2}h_{j}^{1}}\right)
\end{equation}

\begin{equation}
p(h_{m}^{2} = 1 | \textbf{h}^{1}) = \sigma \left( \sum_{j}{W_{im}^{2}h_{i}^{1}} \right)
\end{equation}

\begin{equation}
p(v_{i} = 1 | \textbf{h}^{1}) = \sigma \left( \sum_{j}{W_{ij}^{1}h_{j}}) \right)
\end{equation}

The following section gives a brief overview of the training procedure.

\subsection{Training}

To carry out stochastic gradient ascent on the log-likelihood we make use of the following parameter update rule:

\begin{equation}
\Delta \textbf{W} = \upsilon \left( \mathbb{E}_{P_{data}}\left[ \textbf{vh}^{T}\right] - \mathbb{E}_{P_{model}}\left[ \textbf{vh}^{T}\right] \right) 
\end{equation}

Here, $\mathbb{E}_{P_{data}}\left[ \cdot \right]$ is referred to as the data-dependent expectation while we denote $ \mathbb{E}_{P_{model}}\left[ \cdot \right]$ as the data-independent expectation. In addition, the learning rate $\upsilon$ determines the influence each individual training sample has on the updates of $\textbf{W}$. To train the model's expectations we used stochastic approximation procedures which are outlined in \cite{pmlr-v5-salakhutdinov09a}. 

The data-dependent expectation was approximated using variational learning, where we can characterize the true posterior distribution by a fully factorized distribution \cite[p.1976]{Salakhutdinov:2012:ELP:2330716.2330717}. Besides, the data-independent expectation was approximated using Gibbs sampling \cite[pp.1973-1976]{Salakhutdinov:2012:ELP:2330716.2330717}.

By stacking multiple restricted Boltzmann machines (RBMs), where only two layers are considered simultaneously, the resulting DBM can learn internal representations which enable us to identify complex statistical structures within the hidden layers \cite[p.1970]{Salakhutdinov:2012:ELP:2330716.2330717}. To this end, we adopted the greedy layerwise pre-training which is detailed in \cite{pmlr-v5-salakhutdinov09a}. In this framework \cite{pmlr-v5-salakhutdinov09a} introduce modifications to the first and the last RBM of the stack so that the parameters $\theta = \lbrace \textbf{W}^{1},\textbf{W}^{2} \rbrace$ are initialized to reasonable values. On this basis, the parameters can be improved during the approximate likelihood learning of the entire DBM \cite[p.3175]{hess2017partitioned}.

In our analyses, we set the number of visible nodes $\textbf{v}$ and the number of nodes in the first hidden layer $\textbf{h}^{(1)}$ to seven in order to represent each weekday. We use one node in the terminal hidden layer $\textbf{h}^{(2)}$ since we aimed to detect two groups of usage patterns. Hence, an active node in the terminal hidden layer represents one group while an inactive node represents another pattern. Furthermore, we set the learning rate $\upsilon$ to $0.007$ during pre-training and increased it to $0.008$ for the training of the entire DBM. The number of epochs was held constant at 40 for pre-training as well as for training the entire DBM. Data analysis  was performed using Julia Version 0.6.2 and the Julia package BoltzmannMachines.jl (\emph{https://github.com/binderh/BoltzmannMachines.jl.git}).

\section{Application}

\begin{figure}[H]
	\centering
\includegraphics[width = \textwidth]{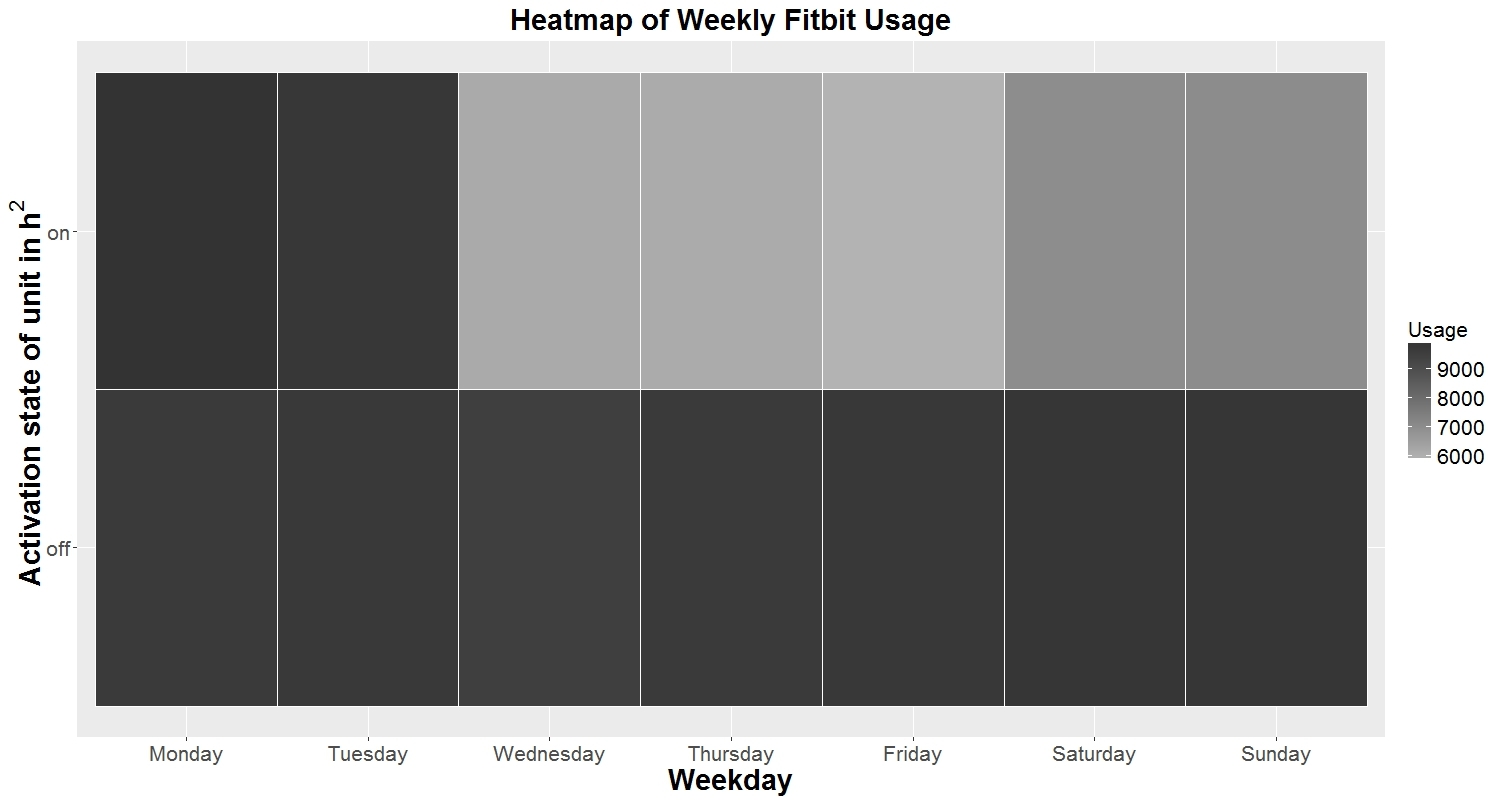}
	\caption{Heatmap of weekly activity tracker usage generated by the DBM. The upper pattern “on” shows a group that frequently uses activity trackers on Mondays and Tuesdays. The lower row indicates a group with regular usage throughout the whole week.}
	\label{fig1}
\end{figure}

After having obtained estimates of the parameters $\theta$, we used the DBM to generate new observations for the visible layer to explore the learned structure. Specifically, we used the DBM to compute the deterministic potential for the activation of the hidden nodes $\textbf{h}^{1}$ given that the nodes in the terminal layer $\textbf{h}^{2}$ were active/inactive. We then propagated the deterministic potential through the network to obtain the visible potential. Subsequently, we generated $10,000$ uniformly distributed random numbers between $0$ and $1$ and assigned the value $1$ if the visible potential was higher than the random number. Next, we used the generated data to plot the learned patterns in a heat map displayed in Figure \ref{fig1}. From the graph, we can distinguish two clear usage patterns. The upper pattern denoted as "on" shows that there is the tendency to use activity trackers at the beginning of the week on Monday and Tuesday and slightly increased usage on the weekend, while the lower pattern indicates a high usage throughout the whole week.

\section{Discussion}

We have presented a deep learning approach, more precisely a deep Boltzmann machine, to model the complex joint distribution of activity tracker data and showed that it can be used successfully to extract meaningful activity tracker usage patterns. Most importantly, we were able to reveal two distinct weekly usage patterns in which one group mostly uses trackers on Mondays and Tuesdays, whereas the other uses trackers during the entire week.
One limitation is our sample size of 29 individuals. Being able to acquire more data and validate the model is an essential next step. Besides, integrating other measurements like heart rate and sleeping behavior could improve our model substantially. Based on this, we could adapt the structure of our DBM to find even more usage patterns or model activity intensity. Another downside regarding our methodology is the dichotomization of step counts which gives away valuable information in the data. This can be addressed by using different cutoffs to incorporate the number of steps per day, but other, direct modeling approaches may be more useful.
Nevertheless, we believe our work could be the basis for future studies which take the continuous measurements from activity trackers into account. To further our research, we plan to use the partitioning approach presented in \cite{hess2017partitioned}. Doing this, we could potentially combine a binary RBM, to model the presence/absence of tracker usage, with a Gaussian RBM, for learning the activity intensity patterns and for modeling the temporal structure over a longer time period.

\newpage

\bibliography{DBM}
\bibliographystyle{plainnat}

\end{document}